\documentclass[10pt,twocolumn,letterpaper]{article}

\usepackage[pagenumbers]{cvpr} 
\usepackage[dvipsnames]{xcolor}
\usepackage{colortbl}
\usepackage{indentfirst}
\usepackage{makecell}
\usepackage{multirow}
\usepackage{pgf}
\usepackage{pifont}
\usepackage{tabularx}
\usepackage{tikz}
\usepackage{float}
\definecolor{cvprblue}{rgb}{0.21,0.49,0.74}
\definecolor{steelblue}{rgb}{0.27,0.51,0.71}
\definecolor{clipscolor}{rgb}{0.906,0.702,0.886}
\definecolor{clipbcolor}{rgb}{0.588,0.871,0.784}
\definecolor{sfcolor}{rgb}{0.694,0.886,0.988}
\definecolor{panncolor}{rgb}{0.961,0.871,0.702}
\definecolor{objcolor}{rgb}{0.65,0.76,0.88} 
\definecolor{dreamcolor}{rgb}{0.471, 0.0, 0.0} 
\definecolor{comcolor}{rgb}{1.0, 0.686, 0.8} 
\usepackage[pagebackref,breaklinks,colorlinks,citecolor=cvprblue]{hyperref}
\usepackage[normalem]{ulem}

\newcommand{\bfsection}[1]{\vspace*{0.05cm}\noindent\textbf{#1.}}

\def\paperID{3379} 
\def\confName{CVPR}
\def\confYear{2025}

\title{Affordance-Aware Object Insertion via Mask-Aware Dual Diffusion}

\author{Jixuan He$^{1,2,}$\thanks{Equal Contribution.}$^{\ ,}$\thanks{Work done while at Harvard University.}\quad Wanhua Li$^{1,}\footnotemark[1]$\quad Ye Liu$^{1,3}$\quad Junsik Kim$^{1}$\quad Donglai Wei$^{4}$\quad Hanspeter Pfister$^{1}$\\
$^{1}$Harvard University\quad$^{2}$ Cornell Tech\quad$^3$The Hong Kong Polytechnic University\quad$^4$ Boston College
}

\newcommand{\humandot}{\raisebox{-0.65pt}{
\begin{tikzpicture}
\fill[clipscolor] (0,0) circle (.8ex);
\draw[black, line width=0.5pt] (0,0) circle (.8ex);
\end{tikzpicture}}}

\newcommand{\gligendot}{\raisebox{-0.65pt}{
\begin{tikzpicture}
\fill[clipbcolor] (0,0) circle (.8ex);
\draw[black, line width=0.5pt] (0,0) circle (.8ex);
\end{tikzpicture}}}

\newcommand{\sddot}{\raisebox{-0.65pt}{
\begin{tikzpicture}
\fill[sfcolor] (0,0) circle (.8ex);
\draw[black, line width=0.5pt] (0,0) circle (.8ex);
\end{tikzpicture}}}

\newcommand{\pbedot}{\raisebox{-0.65pt}{
\begin{tikzpicture}
\fill[panncolor] (0,0) circle (.8ex);
\draw[black, line width=0.5pt] (0,0) circle (.8ex);
\end{tikzpicture}}}

\newcommand{\objdot}{\raisebox{-0.65pt}{
\begin{tikzpicture}
\fill[objcolor] (0,0) circle (.8ex);
\draw[black, line width=0.5pt] (0,0) circle (.8ex);
\end{tikzpicture}}}

\newcommand{\dreamdot}{\raisebox{-0.65pt}{
\begin{tikzpicture}
\fill[dreamcolor] (0,0) circle (.8ex);
\draw[black, line width=0.5pt] (0,0) circle (.8ex);
\end{tikzpicture}}}

\newcommand{\comdot}{\raisebox{-0.65pt}{
\begin{tikzpicture}
\fill[comcolor] (0,0) circle (.8ex);
\draw[black, line width=0.5pt] (0,0) circle (.8ex);
\end{tikzpicture}}}

\begin{document}
\maketitle
\begin{abstract}
As a common image editing operation, image composition involves integrating foreground objects into background scenes. In this paper, we expand the application of the concept of \textit{Affordance} from human-centered image composition tasks to a more general object-scene composition framework, addressing the complex interplay between foreground objects and background scenes. Following the principle of \textit{Affordance}, we define the affordance-aware object insertion task, which aims to seamlessly insert any object into any scene with various position prompts. To address the limited data issue and incorporate this task, we constructed the SAM-FB dataset, which contains over 3 million examples across more than 3,000 object categories. Furthermore, we introduce a strong baseline: a \textbf{M}ask-\textbf{A}ware \textbf{D}ual \textbf{D}iffusion (MADD) model, which utilizes a dual-stream architecture to simultaneously denoise the RGB image and the insertion mask. By explicitly modeling the insertion mask in the diffusion process, MADD effectively facilitates the notion of affordance.  Experimental results show that our method outperforms the state-of-the-art methods and exhibits strong generalization performance on in-the-wild images. Please refer to our code on \href{https://github.com/KaKituken/affordance-aware-any}{https://github.com/KaKituken/affordance-aware-any}.
\end{abstract}
    
\section{Introduction}
\label{sec:intro}

In the image composition scenario, common sense guides our perception of the authenticity of synthesized images: a person cannot levitate, a water bottle needs a surface for support, and a boat should be floating in the water rather than on the ground. Deviation from such common sense often leads to semantic inconsistencies on synthesized images. From a composition perspective, the background's semantic richness plays a pivotal role in defining the placement and characteristics of foreground objects. To better describe the influence of background semantics on the foreground, we borrow the concept of \textit{affordance} into object-scene composition tasks. Previously, Kulal \etal~\cite{kulal2023putting} explored the concept of human affordance for image synthesis. Their work, focused on inserting humans into masked scenes, extends beyond mere color or view adjustments. To generalize the setting to arbitrary object-scene synthesis, we term the new task \textit{Affordance-Aware Object Insertion}. This task challenges models to identify suitable locations and make necessary adjustments to foreground objects, ensuring the generated images adhere to physical laws.

This work aims to build a foundation model for affordance-aware object insertion, which can put any object into any scene as shown in Figure \ref{fig:teaser}. There are three primary challenges involved. First, the model must accurately recognize the appropriate affordance relationship between a background and the foreground object to be inserted. Adjustments to the inserted foreground object is crucial for achieving the intended semantic consistency.
The second challenge lies in the model’s ability to generalize across a diverse range of foreground objects. Previous generative image editing methods, such as textual inversion~\cite{gal2022image}, DreamBooth~\cite{ruiz2023dreambooth}, and DreamEdit~\cite{li2023dreamedit}, are subject-specific thus hard to generalize. Our goal is to train a model that can generalize to any object.
Third, our model should support a variety of input prompts for users to specify insertion locations, ranging from sparse formats like points and bounding boxes to dense masks. Moreover, in the absence of explicit prompts, the model can autonomously determine appropriate insertion locations by analyzing the semantic content of both the background and foreground.

We address these challenges through three components: \textbf{task}, \textbf{data}, and \textbf{model}. Extending the concept of \textit{affordance} beyond Kulal \etal's initial scope~\cite{kulal2023putting}, the task of affordance-aware object insertion aims to place an arbitrary object into any scene, accommodating different positional prompts, even in the absence of explicit positional cues. It has significant implications for applications such as automated dataset synthesis. To support this task, a large-scale dataset is necessary. Existing image composition datasets, such as DreamEditBench~\cite{li2023dreamedit}, are limited in terms of the diversity of foreground object categories and the number of training samples. 
To overcome these limitations, we curate a new dataset called SAM-FB which is derived from SA-1B~\cite{kirillov2023segment} for affordance learning. SAM-FB contains a variety of foreground object categories and over 3 million samples. With SAM-FB, we further introduce a strong baseline: a \textbf{M}ask-\textbf{A}ware \textbf{D}ual \textbf{D}iffusion (MADD) model to utilize the large-scale data, which is a diffusion-based framework that facilitates the seamless integration of diverse objects into any scene.  During the denoising procedure, object position is progressively refined while the target RGB image is synthesized simultaneously, ensuring accurate alignment between objects and positions to achieve affordance-aware insertion. 
Furthermore, we 
present a unified representation for sparse and dense prompts,
enabling our model to effectively respond to various types of position inputs.

The contributions of this work are as follows: 
\begin{itemize}
\item We introduce affordance-aware object insertion, extending object-scene composition with affordance guidance to enable realistic insertions across diverse prompts.
\item We present SAM-FB, a large-scale dataset with over 3 million samples spanning diverse object categories to support affordance-aware insertion.
\item We propose a strong baseline model with a dual-stream architecture that denoises object appearance and the insertion mask, facilitating affordance learning.
\end{itemize}
\section{Related Work}
\label{sec:related}
\bfsection{Affordance}
J.J. Gibson~\cite{bornstein1980ecological} first introduces the concept of affordance, then a series of papers~\cite{brooks2022hallucinating,fouhey2015defense,gupta20113d,li2019putting} dug into this concept and brought it into the image synthesis. Initially grounded in psychology, the concept emphasizes that an object's appearance should correspond with its utilitarian aspects as perceived by humans. 
Further exploration within the field of image synthesis involved adjusting the orientation and gestures of generated human figures to align with their background.
Therefore, prior work primarily focuses on the interaction between humans and objects~\cite{gkioxari2018detecting,yao2010modeling,zhu2014reasoning} or the human-scene relationship~\cite{fouhey2012people,delaitre2012scene,wang2017binge}. Kulal~\etal~\cite{kulal2023putting} made progress by introducing a model trained on person movement video data for placing people within scenes and adjusting its pose according to the surroundings. However, their model’s scope was limited to human figures. Despite these advancements, the concept of "affordance" in image composition, which encompasses the positioning, viewing angle, and color harmony of objects within scenes, has remained relatively unexplored.
Our work offers a generalized and versatile solution for object-scene composition.

\begin{table*}[!htp]
\setlength{\tabcolsep}{1mm}
\centering
    \begin{tabular}{lccccc}
    \toprule
     Model & Task & Dataset & Sample & Category & Affordance Consistency \\
    \midrule
    \dreamdot~\cite{li2023dreamedit} &Customized Image Composition& DreamEditBench & 440 & 22 & High (Real Paired) \\
    \comdot~\cite{lu2023dreamcom}&Customized Image Composition & MureCom & 640 & 32 & High (Real Paired)\\
     \pbedot~\cite{yang2023paint}, \objdot~\cite{song2023objectstitch} & Image-guidance Object Insertion & Open-Images-v4 & 1.9M & 600 & Low (Masked) \\
     \gligendot~\cite{li2023gligen} & Image-guidance Object Insertion & COCO2014 & 165K & 80 & Low (Masked)\\
    \humandot~\cite{kulal2023putting} & Human Affordance Insertion & Private& 2.4M & 1 & High (Video) \\
    \midrule
    \textbf{MADD (ours)} & Object Affordance Insertion &SAM-FB  & \textbf{3.2M} & \textbf{3,439} & High (Inpainted) \\
    \bottomrule
    \end{tabular}
    \caption{Dataset choice and comparison. We inspect various potential datasets for affordance insertion. SAM-FB contains significantly more samples and object categories.
    \protect\dreamdot~\hspace{0.1mm} DreamEdit, 
    \protect\comdot~\hspace{0.1mm} DreamCom,
    \protect\pbedot~\hspace{0.1mm} PBE, \protect\objdot~\hspace{0.1mm} ObjectStitch,
    \protect\gligendot~\hspace{0.1mm}GLIGEN, 
    \protect\humandot~\hspace{0.1mm} Human Affordance. }
    \label{table:dataset}
    \vspace{-3mm}
\end{table*}

\bfsection{Image Editing}
Image editing aims to modify existing images using generative models. Generally, image editing can be divided into semantic editing (\eg adding or removing objects, changing the background), style editing (\eg altering color or texture), and structural editing (\eg changing object size or viewpoint). By utilizing generative models like GANs~\cite{goodfellow2014generative} or Diffusion~\cite{dhariwal2021diffusion}, users can edit image content by providing instructions and multi-modal prompts. For instance, InstructPix2Pix~\cite{brooks2023instructpix2pix} and MoEController~\cite{li2023moecontroller} allow semantic and style editing through text-based instructions provided by users, while Imagen Editor~\cite{wang2023imagen} enables more precise control of edit locations by accepting user-provided masks for the areas to be edited. Additionally, other image editing models, such as TF-ICON~\cite{lu2023tf} and ImageBrush~\cite{yang2024imagebrush}, accept reference images from users, constraining the appearance of objects during editing. 

\bfsection{Image Composition}
Image composition is a sub-task within image editing, primarily focused on controllably inserting foreground objects into a background based on reference images. Recently, the utilization of generative models empowers high-quality and controllable image composition. Diffusion models have been successfully applied in various domains, including image composition~\cite{song2023objectstitch,liu2023hyperhuman}, video generation~\cite{ho2022video}, and data augmentation~\cite{trabucco2023effective}. Stable Diffusion (SD)~\cite{rombach2022high} introduced methods for blending feature and semantic information from images and text, enabling object generation based on prompt instructions. Inpainting~\cite{bertalmio2000image} is a common approach to achieve image composition. SD-based inpainting models~\cite{yu2018generative,lugmayr2022repaint,li2022mat} allow users to repaint a specific region with a mask, but it's hard to insert a desired object into that region. Blended Diffusion~\cite{avrahami2022blended} enables finer control by leveraging text descriptions of the foreground, but it's crucial for user to describe the foreground precisely. PBE~\cite{yang2023paint},  ObjectStitch~\cite{song2023objectstitch}, and GLI-GEN~\cite{li2023gligen} utilize reference images to incorporate rich visual information, achieving better perspective and color harmony. However, these models require users to provide precise positional cues, such as bounding boxes or masks, to indicate the insertion location. Our proposed method offers a viable solution for handling vague positional information, such as point prompts or even blank prompts.

\section{Dataset}


\label{sec:dataset}
\begin{figure*}[t]
  \centering
   \includegraphics[width=1.0\linewidth]{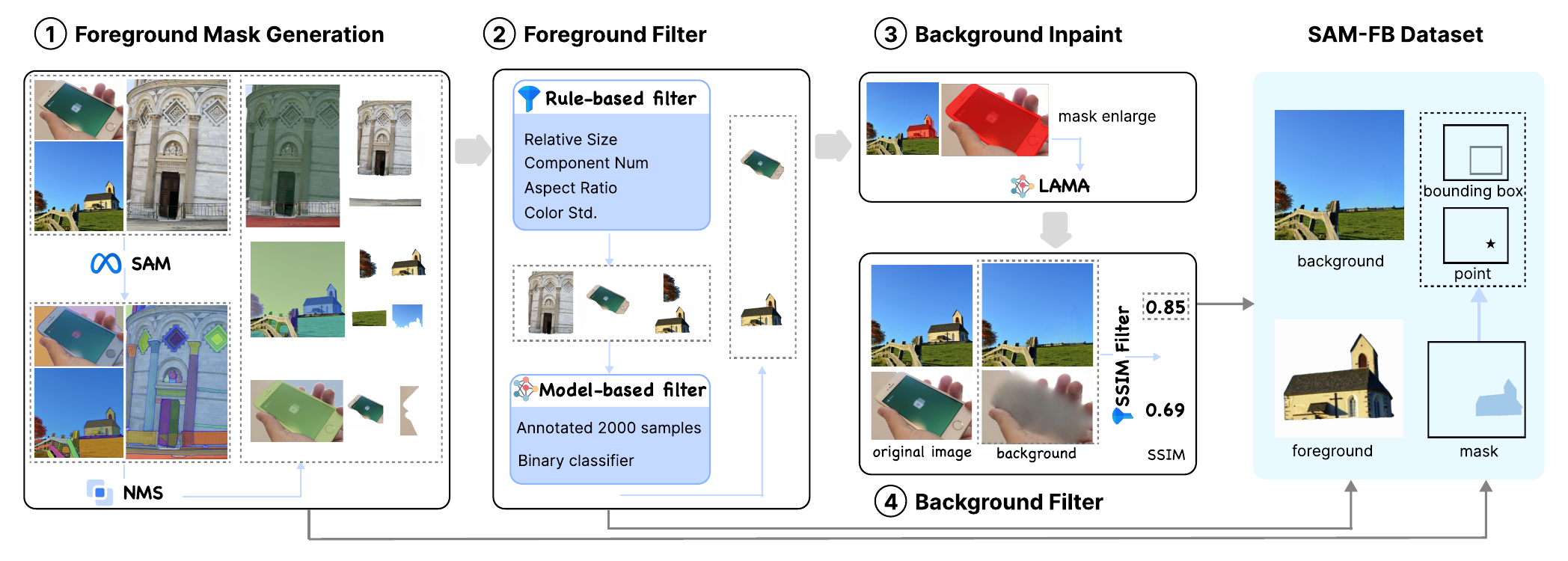}
   \caption{Dataset construction pipeline for SAM-FB. The pipeline automatically converts any input image into a tetrad output through four stages, ensuring high-quality foreground and background retention via a rigorous data quality control process.}
   \label{fig:pipeline}
   \vspace{-5mm}
\end{figure*}

Instead of manually designing affordance, we use a data-driven approach to extract affordance relationships between the object and background from the large-scale dataset.

\subsection{Motivation \& Design Decisions}
To support affordance-aware object insertion across various objects and scenes, the dataset must provide rich contextual information and extensive coverage and preserve natural affordance cues between objects and their surroundings.
Therefore, we designed the dataset with the following key characteristics:
1) Well-aligned input-output pairs. Rather than using techniques like DreamBooth~\cite{ruiz2023dreambooth} to memorize each object individually, we adopted supervised fine-tuning to directly learn from large-scale object distributions. Consequently, well-aligned input-output pairs, represented as $(\boldsymbol{f},\boldsymbol{b},\boldsymbol{p},\boldsymbol{x})$ tuples, are essential. Here, $\boldsymbol{f}$, $\boldsymbol{b}$, $\boldsymbol{p}$, and $\boldsymbol{x}$ denote the foreground image, background image, position prompt, and ground truth image, respectively.
2) Sufficient training samples. Training a model capable of generalizing to diverse scenarios requires a substantial number of training samples.
3) A wide variety of foreground objects. The dataset must encompass a diverse range of foreground objects to ensure that the trained model can generalize across different object categories.
However, existing datasets either lack sufficient object diversity or introduce artifacts that disrupt these affordance relationships.

The $(\boldsymbol{f},\boldsymbol{b},\boldsymbol{p},\boldsymbol{x})$-formatted data can be derived from multiple datasets relevant to this task, as shown in Table~\ref{table:dataset}. One option is to use datasets designed for Customized Image Composition. Datasets such as DreamEditBench~\cite{li2023dreamedit} and MureCom~\cite{lu2023dreamcom} are manually captured by placing the same foreground object in different backgrounds under controlled camera settings. However, the cost of constructing such high-quality paired datasets is substantial, resulting in relatively small dataset sizes.
Referring to the Image-Guided Object Insertion task, another approach is to leverage large-scale object detection datasets such as COCO~\cite{lin2014microsoft} and Open-Images~\cite{kuznetsova2020open} to simulate foreground-background pairs. This is achieved by using bounding boxes as position prompts and extracting foreground objects from the original image using masks, treating the remaining portion as the background. However, these datasets are still insufficient to guarantee robust generalization, particularly due to their limited object category diversity. Moreover, masking out the foreground leaves an opaque region in the background, which disrupts affordance relationships. For HumanAffordance~\cite{kulal2023putting}, they use a vast number of human-centered video to capture the affordance relationship, but the category is human only and the dataset is private.
To address these limitations and obtain a sufficient number of training samples while ensuring diversity in foreground objects, we turned to the SA-1B dataset~\cite{kirillov2023segment}, which consists of 11 million high-quality images collected from various real-world scenes and 1 billion well-annotated class-agnostic masks.
Based on SA-1B, we constructed the SAM-FB benchmark dataset for the affordance-aware object insertion task.

\subsection{Dataset Construction Pipeline}

Figure~\ref{fig:pipeline} illustrates the pipeline used to construct the aforementioned $(\boldsymbol{f},\boldsymbol{b},\boldsymbol{p},\boldsymbol{x})$ formatted data. The pipeline automatically transforms any input image into a tetrad output. To enhance affordance representation and improve the overall quality of the SAM-FB dataset, the dataset construction process consists of four key stages.

\bfsection{Foreground Mask Generation} 
For each unannotated input image, we first apply the Segment Anything Model (SAM) to generate object masks as candidates. Since SAM produces masks at different levels of granularity with varying confidence scores, we perform Non-Maximum Suppression (NMS) with a high threshold (0.6) to remove redundant masks and filter out sub-object-level masks. The remaining masks are used to extract candidate foreground objects.

\begin{table}[t]
\centering
\resizebox{0.85\linewidth}{!}{  
    \begin{tabular}{lccc}
    \hline
    Filter condition & Threshold & Reserved Percentage \\
    \hline
    None (Initial) & -- & 100\% \\
    Relative Size & [0.1, 0.75] & 7.10\% \\
    Aspect Ratio & $\leq$ 3 & 6.88\% \\
    Components Num. & $\leq$ 4 & 6.71\% \\
    Color Std. & $\geq$ 45 & 1.69\% \\
    ResNet50 Score & $\geq $ 0.7 & 0.25\%\\
    \hline
    \end{tabular}
    }
    \caption{Reserved percentages for foreground quality control filters. We apply a combination of rule-based and learning-based conditions to ensure high-quality foreground objects are retained through a rigorous filtering process.}
    \label{table:quality}
    \vspace{-3mm}
\end{table}

\bfsection{Foreground Filtering}
Although NMS removes some low-quality masks, SAM segmentation can still produce unintended elements due to challenges in controlling granularity and semantic consistency. The resulting masks often suffer from three main issues:
Incompleteness: Some objects are partially occluded, resulting in fragmented masks that only capture small portions of the original object (e.g., half of a tree);
Background Masks: Some masks mistakenly segment background elements (e.g., sky, grassland);
Undersized / Oversized Objects: Very small objects often have blurry details, reducing dataset quality. To enhance foreground quality, we employ a hybrid quality control (QC) approach combining rule-based and learning-based filtering. 

For rule-based filtering, we remove low-quality masks with four constraints: relative size (too small/large objects), aspect ratio (overly thin objects), component count (disjointed objects), and color standard deviation (pure-color backgrounds). While rule-based filtering effectively removes outliers, ensuring object completeness is challenging using predefined rules alone. To address this, we manually annotate 2,000 foreground images with binary labels (\texttt{good} vs. \texttt{bad}) and fine-tune a ResNet-50~\cite{he2016deep} classifier for binary classification. The trained classifier assigns a quality score to each foreground object, further refining the dataset. Table \ref{table:quality} demonstrates the specific threshold for each filter operation of our data quality control stage. With these filter operations, only 0.25\% of the masks are left, which ensures the quality of the obtained dataset. 

\bfsection{Background Inpainting}
High-quality foregrounds are then cropped from the original images. Instead of covering the removed regions with opaque masks, which act as strong artificial cues for object position and size, we employ LAMA~\cite{suvorov2021resolution} for background inpainting. LAMA is a simple yet effective method for object removal and hole-filling. Following GroundingSAM~\cite{liu2023grounding,ren2024grounded}, we expand the object mask boundary slightly to prevent residual foreground content from remaining in the background. By eliminating such visual cues, our dataset encourages models to infer context-aware object placement based on scene affordances.

\bfsection{Background Inpainting}
Despite careful inpainting, some regions inevitably contain inpainting artifacts. To assess the quality of inpainted backgrounds, we compute the Structural Similarity Index (SSIM) between the inpainted background and the original image. We discard any inpainted background with an SSIM score below a certain threshold to maintain high visual consistency. We empirically set the SSIM threshold to 0.8 based on preliminary experiments assessing inpainting realism. If a background fails the SSIM filter, we discard the entire foreground-background pair, even if the foreground quality is high. This ensures that only high-quality samples are included in the dataset.

To evaluate quality improvements, we compute the Inception Score (IS) for foregrounds and Fréchet Inception Distance (FID) for backgrounds, comparing values before and after quality control (QC). After QC, foreground IS improves from 9.55 to 16.68, indicating significantly enhanced foreground clarity, while background FID decreases from 9.23 to 8.46, reflecting better distribution consistency, similar to inpainting tasks. Since FID requires a reference distribution, it is less meaningful for foregrounds, whereas IS is not a reliable metric for complex backgrounds.
To justify the diversity of our proposed dataset, we use RAM~\cite{zhang2023recognize} to recognize the categories of foregrounds we keep. We have identified 3439 different categories, which is a vast improvement over previous datasets. There are no manual annotations required in the pipeline, and users can easily scale up the dataset further with any new source images easily.

\section{Method}
\label{sec:method}



\subsection{Mask-Aware Dual Diffusion Model (MADD)}
\bfsection{Preliminaries}
Given a foreground object $\boldsymbol{f}$, a background scene $\boldsymbol{b}$, and a position prompt $\boldsymbol{p}$, the goal is to synthesize an image $\boldsymbol{x}$ that integrates the object into the scene, while adhering to affordances and aligning with the position prompt.

\begin{figure}[t]
  \centering
   \includegraphics[width=\linewidth]{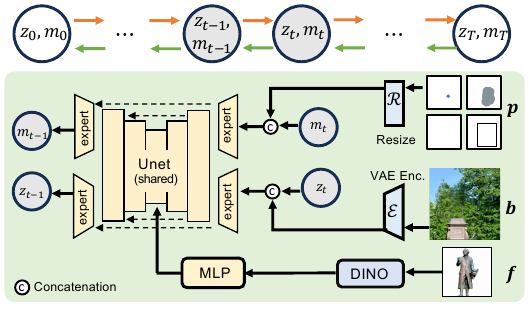}
   \caption{Mask-aware Dual Diffusion Model (MADD). 
   The RGB image feature $\mathbf{z}$ and object mask $\mathbf{m}$ are jointly denoised, conditioning on the embeddings of the foreground object $\mathbf{f}$, background object $\mathbf{b}$, and the prompt $\mathbf{p}$. (green: reverse process $t\rightarrow t\mathrm{-}1$)}
   \label{fig:dual}
   \vspace{-3mm}
\end{figure}

To develop an effective diffusion model, we first extract image features $\boldsymbol{z}=\mathcal{E}(x)$ using a pre-trained encoder.
Then we predict the object insertion mask $\boldsymbol{m}$ capturing object's location and size, which can be directly derived from the difference between $\boldsymbol{x}$ and $\boldsymbol{b}$ in the training data.
Our MADD model $\mathcal{G}$ builds upon the latent structural diffusion framework~\cite{liu2023hyperhuman}, simultaneously denoising $\boldsymbol{\hat{z}}$ and $\boldsymbol{\hat{m}}$ (Figure~\ref{fig:dual}): 
\begin{equation}
    \hat{\boldsymbol{z}},\hat{\boldsymbol{m}} = \mathcal G(\boldsymbol{f},\boldsymbol{b},\boldsymbol{p}),  
\end{equation}
where the estimated image is reconstructed by $\boldsymbol{\hat{x}}=\mathcal{D}(\boldsymbol{\hat{z}})$ with the corresponding image decoder.

\bfsection{Forward Process}
The noised feature maps are
\begin{equation}    
\boldsymbol{z_{t_z}} = \alpha_{t_z} \mathbf z + \sigma_{t_z} \boldsymbol\epsilon_{t_z},~~~
\boldsymbol{m_{t_m}} = \alpha_{t_m} \mathbf m + \sigma_{t_m} \boldsymbol\epsilon_{t_m},
\end{equation}
where $\epsilon_z$ and $\epsilon_m$ $\sim\mathbf{N}(0,I)$ are three independently sampled Gaussian noise, and the time step $t_z$ and $t_m$ are sampled from a uniform distribution $\mathcal U[0,1]$.

\bfsection{Reverse Process}
The denoiser $\hat{\boldsymbol\epsilon}_\theta$ simultaneously denoises both $\boldsymbol{z_{t_z}}$ and $\boldsymbol{m_{t_m}}$, which can be trained with:
{\small
\begin{equation}
\mathcal L=\mathbb E\left[\|\boldsymbol\epsilon_z-\hat{\boldsymbol\epsilon}_\theta(\mathbf z_{t_z}; \mathbf{f},\mathbf{b},\mathbf{p}))\|^2\right]+
\mathbb E\left[\|\boldsymbol\epsilon_m-\hat{\boldsymbol\epsilon}_\theta(\mathbf{m_{t_m}};\mathbf{f},\mathbf{b},\mathbf{p})\|^2\right].\nonumber
\end{equation}
}

The denoiser model consists of the UNet model and the three encoders for the foreground object $\boldsymbol{f}$, the background scene $\boldsymbol{b}$, and the position prompt $\boldsymbol{p}$, respectively.

\subsection{Architecture Details}

\bfsection{UNet Model}
 We utilize a single UNet with two expert input-output branches to denoise them simultaneously. 
The two tasks share the entire UNet except for the  \verb|conv_in|, first \verb|Down Block|, last \verb|Up Block|, and the \verb|conv_out|. These independent blocks serve as the Expert input-output branch. Skip connection is also performed between the corresponding Expert input and output branches. 

\bfsection{Foreground Encoder}
In MADD, foreground images serve as a guidance condition and are incorporated into the model via cross-attention. The foreground embedding is extracted using a ViT-like visual encoder. While SD employs the CLIP~\cite{radford2021learning} encoder for text-aligned semantics, we follow Efficient-3DiM~\cite{jiang2023efficient} and use DINOv2~\cite{oquab2023dinov2} to preserve fine-grained object details, making it more suitable for our task.


\bfsection{Background Encoder} Following SD-based image editing methods~\cite{song2023objectstitch, brooks2023instructpix2pix}, we employ a pre-trained VAE encoder to convert the background image into a 4-channel latent representation. Since the background is pixel-aligned with the output, we concatenate its latent map with $\mathbf{z_t}$ before feeding them into the U-Net. The pre-trained VAE decoder then reconstructs the final RGB image from the latent output.

\bfsection{Position Prompt Encoder} 
We design a unified representation to process different position prompts. Sparse and dense prompts are converted into a 1-channel position map, guiding foreground object placement. Point prompts are transformed into Gaussian heatmaps, bounding boxes are filled with ones inside and zeros outside, and masks are binary maps. For null prompts, an all-one image implies all positions are possible. The position map is resized to match the latent map's spatial dimensions and concatenated with it before being fed into the diffusion U-Net.

\begin{figure*}[t]
  \centering
   \includegraphics[width=\linewidth]{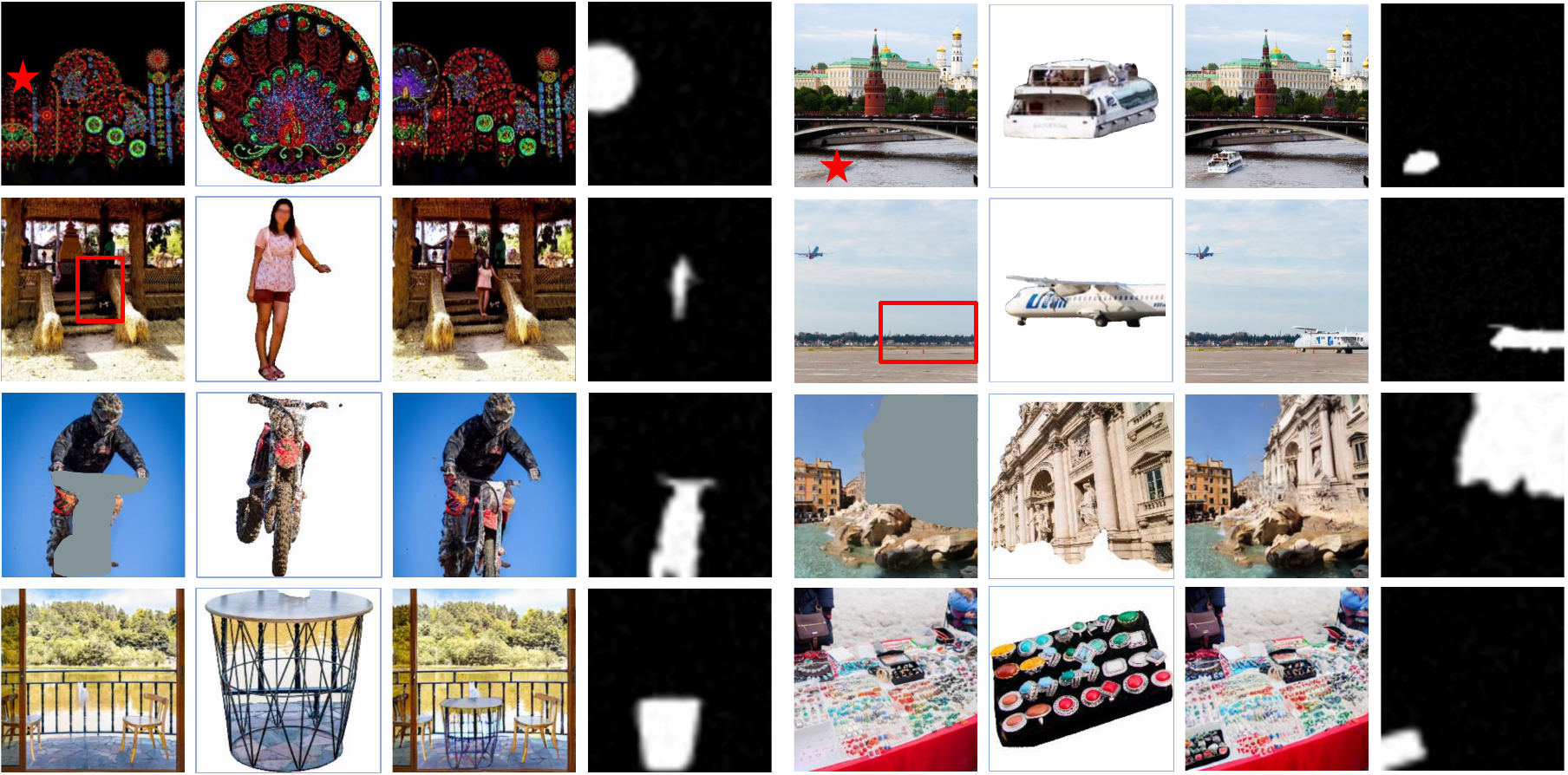}
   \caption{Qualitative results of MADD on the SAM-FB test set. Each row corresponds to one type of prompt, \ie, point, bounding box, mask, and null, respectively. Our MADD simultaneously predicts the RGB image and the object mask.}
   \label{fig:sam-fb}
   \vspace{-3mm}
\end{figure*}

\subsection{Implementation Details} 
We initialize our model with a pre-trained Stable Diffusion Inpainting model and apply strong data augmentation to enhance affordance learning. To optimize fine-tuning, we use a coarse-to-fine strategy: training at \(128\times128\) resolution (batch size 1024) for 35K steps, then fine-tuning at \(256\times256\) (batch size 256) for 15K steps. 
\section{Experiments}

\subsection{Results on the SAM-FB Test Set}

\begin{table}[t]
\centering
\resizebox{0.85\linewidth}{!}{  
\setlength{\tabcolsep}{4pt}
    \begin{tabular}{lcccccc}
    \toprule
    \multirow{2}{*}{\textbf{Method}} & \multicolumn{2}{c}{\textbf{FID}$\downarrow$}  & \multicolumn{2}{c}{\textbf{CLIP Score}$\uparrow$} & \multicolumn{2}{c}{\textbf{MSE}$\downarrow$}\\
    \cline{2-7} 
    & mask & bbox & mask & bbox & mask & bbox \\
    \midrule
    \sddot~\cite{rombach2022high} & 15.41 & 15.47 &  0.7079 & 0.8058 &860 & 883\\
    \pbedot~\cite{yang2023paint} & 33.68 & 24.59 & -- & 0.7664 & 2373 & 1615\\
    \gligendot~\cite{li2023gligen} & - & \uline{13.82} & -- & 0.7896 & -- & 817\\
    \humandot~\cite{kulal2023putting} & 14.49 & 14.42 & 0.8014 & \uline{0.8637} & 857 & 845 \\
    \midrule
    \textbf{Ours} & \textbf{13.53} & \uline{13.60} & \textbf{0.8727} & \uline{0.8658} & \textbf{760} & \textbf{775} \\
    \bottomrule
    \end{tabular}}
    \vspace{-2mm}
    \caption{Method comparisons on the SAM-FB test set. \protect\sddot~\hspace{0.1mm} Stable Diffusion, \protect\pbedot~\hspace{0.1mm} PBE, \protect\gligendot~\hspace{0.1mm}GLIGEN, \protect\humandot~\hspace{0.1mm} Human Affordance. }
    \label{table:quan}
\vspace{-3mm}
\end{table}

\bfsection{Evaluation Metrics} 
Following the previous work~\cite{lu2023tf, jeong2023zero, kawar2023imagic, couairon2022diffedit}, we employ the FID score to measure the quality of images obtained by generative models, MSE for pixel similarity and the CLIP score to evaluate the semantic similarity between the edited region and the reference foreground.

\bfsection{Quantitative Results} 
We compare our method with Stable Diffusion~\cite{rombach2022high}, PBE~\cite{yang2023paint}, GLI-GEN~\cite{li2023gligen}, and Human Affordance~\cite{kulal2023putting} on the SAM-FB test set to show the effectiveness of our method. 
For fairness, we fine-tuned GLI-GEN [28] on SAM-FB training data for 2k steps with captions generated by ViT-GPT2~\cite{radford2019language,dosovitskiy2020image} and use Box + Text + Image Inpainting mode to adapt it. Since Human Affordance~\cite{kulal2023putting} does not release training data or model weights, we re-trained their model on SAM-FB.
The results in Table~\ref{table:quan} show that our method attains the best FID score and the highest CLIP score, illustrating our model can be a simple yet strong baseline for affordance-aware object insertion task. Table~\ref{table:prompt} shows that the mask prompt achieves the best results as it provides more accurate position information. 




\begin{table}[t]
\setlength{\tabcolsep}{1.6mm}
\centering
\resizebox{0.85\linewidth}{!}{  
    \begin{tabular}{lccccc}
    \toprule
    \textbf{Prompt} & Mask & Bbox & Point & Null & \textbf{Avg.}\\
    \midrule
    FID & \textbf{13.53}& 13.60 & 13.66&13.96 & 13.69\\
    MSE & \textbf{760} & 775&772 & 860 & 792 \\
    CLIP Score & \textbf{0.8727}& 0.8658 & 0.8567 & 0.8034 & 0.8497 \\
    \bottomrule
    \end{tabular}
    }    
    \vspace{-1mm}
    \caption{Comparison of position prompts on the SAM-FB test set.}
    \label{table:prompt}
    \vspace{-1mm}
\end{table}

\begin{figure*}[t]
   \centering
   \includegraphics[width=\linewidth]{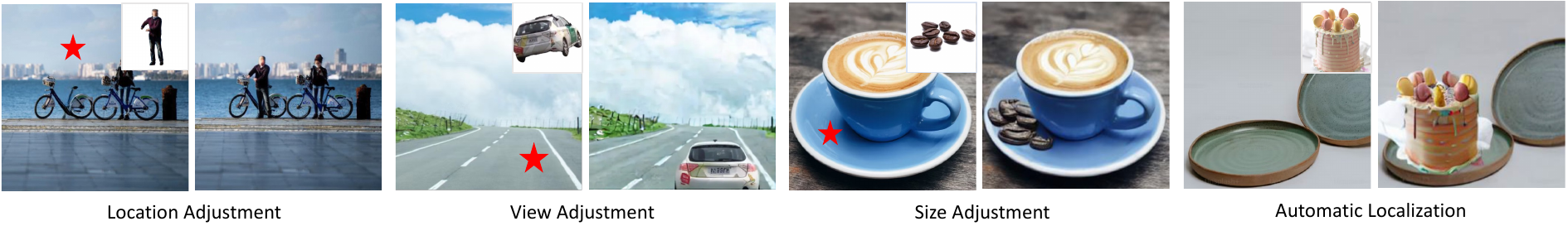}
   \vspace{-3mm}
   \caption{We test ambiguous prompts (points and blank) on in-the-wild images. With point prompts, our model adjusts foreground properties for affordance-aware insertion, while it autonomously finds suitable positions when no prompt is given.}
   \label{fig:affordance}
   \vspace{-4mm}
\end{figure*}
\bfsection{Qualitative Results} 
We left 3786 images as test split.
Figure~\ref{fig:sam-fb} presents the visualization results on the SAM-FB test set. In each group, the leftmost image depicts the background marked with a position prompt.
Our MADD predicts the RGB image and mask of the inserted object, which are shown in the last two images of each group. 
These results demonstrate that MADD not only inserts objects with high quality but also accurately predicts object masks.
\begin{table}[t]
\centering
\resizebox{0.5\linewidth}{!}{  
\begin{tabular}{l|cc}
\hline
Prompt& Non-Null & Null\\
\hline 
IoU ($\uparrow$) & 0.5405 & 0.4696\\
\hline
\end{tabular}
}
\vspace{-2mm}
\caption{Mask evaluation}\label{tab:mask}
\vspace{-3mm}
\end{table}

\begin{table}[t]
\setlength{\tabcolsep}{3mm} 
\centering
\resizebox{0.7\linewidth}{!}{  
\begin{tabular}{lcc}
\toprule
\multicolumn{3}{c}{Model Components} \\
\midrule
Method & FID ($\downarrow$) & CLIP ($\uparrow$) \\
\midrule
Baseline & 18.32  & 0.79 \\
+ Dual diffusion & 14.43 & 0.84 \\
+ Expert branch & \textbf{13.69} & \textbf{0.85} \\
\midrule
\multicolumn{3}{c}{Higher Resolution} \\
\midrule
256 $\times$ 256 & 13.69 & \textbf{0.85} \\
512 $\times$ 512 & \textbf{10.66} & 0.82 \\
\bottomrule
\end{tabular}
\vspace{-2mm}
}
\caption{Ablation study results on SAM-FB test set.}
\label{table:ablation}\vspace{-5mm}
\end{table}

\bfsection{Mask Evaluation} Table~\ref{tab:mask} shows the IoU of the mask between the predicted mask and the ground truth on the SAM-FB test set. 
The relatively high IoU shows that our method learns effective affordance information.

\bfsection{Computational complexity}
We access the computational complexity for the newly added modules including Dual Diffusion and Expert Branch. Using Human Affordance model with DINOv2 as baseline, it has 863.02M parameters with 28.3B FLOPs. Our Expert branches shared most of the parameters with the original baseline model and end up with  873.75M parameters with 31.1B FLOPs.

\begin{figure*}[t]
  \centering
    \includegraphics[width=\linewidth]{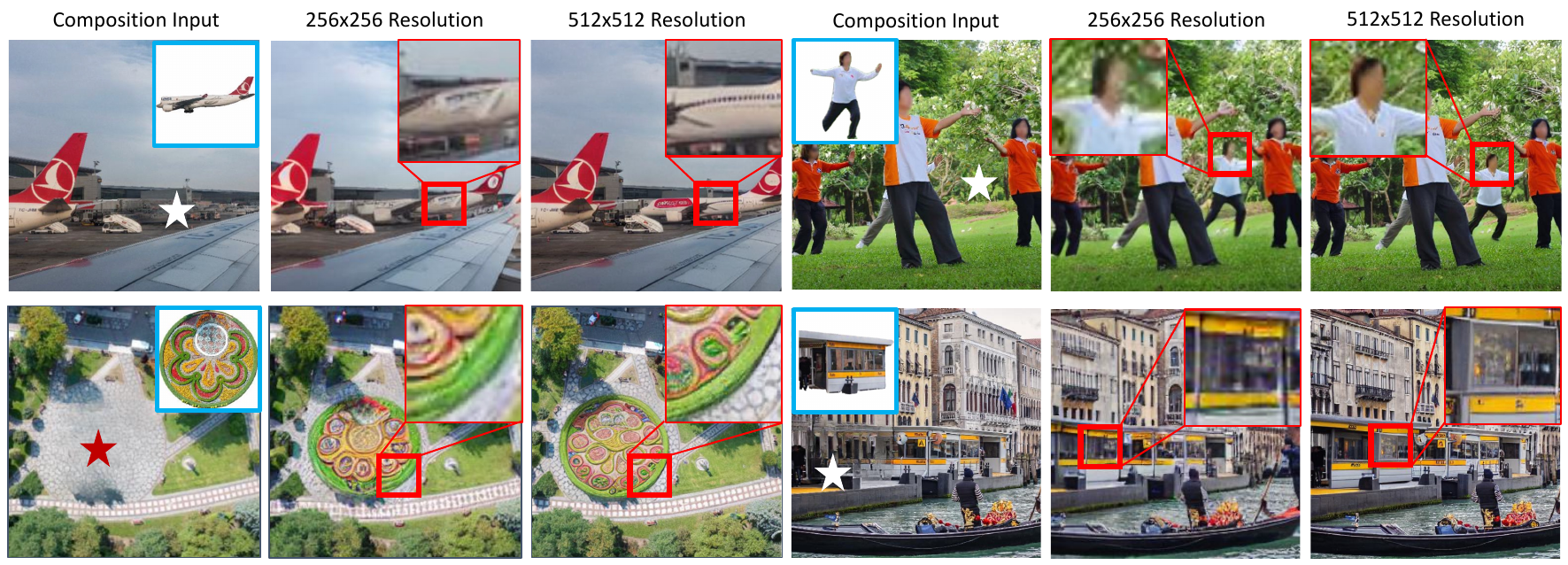}
  \caption{MADD can work on images of higher resolution, generating sharper edges, clearer reflections, improved texture details.}
  \label{fig:higher}
  \vspace{-4mm}
\end{figure*}

\subsection{Ablation Studies}
\bfsection{Model Components} 
Table~\ref{table:ablation} presents the results of our ablation study. For model design, we replace the image encoder in the Human Affordance model with DINOv2 as the baseline. Next, we diffuse RGB image and object mask simultaneously, but sharing the entire UNet. Finally, we use two Expert branches for mask and RGB streams. The results show that all of them improved performance. Metrics are averaged across different position prompts. Please refer to the supplementary for results on each prompt type.

\bfsection{Higher Resolution}
Even though we trained our model at a resolution of 256x256, it is also capable of handling higher-resolution inputs, such as 512x512. To prove that, we then fine-tuned our model on SAM-FB at a resolution of 512x512 for only 200 steps and compared it with the original checkpoint on the test split. Figure~\ref{fig:higher} shows the results of the original checkpoint and the fine-tuned version at higher resolution using the same input. We highlighted the details of the inserted object. It is clearly demonstrated that the model generates sharper edges, clearer reflections, and improved texture details after fine-tuning at higher resolution. The CLIP score drops slightly at 512×512, likely due to the increased complexity of higher-resolution foregrounds, making fine-grained detail synthesis more challenging.

\subsection{Results on In-the-wild Images}
\noindent We evaluate our model on in-the-wild web-crawled images.
\bfsection{Ambiguous Prompts} 
Training with position augmentations enables the model to capture affordance relationships between objects and scenes. MADD refines object position, size, and view for coherence with the background.

\begin{figure}[t]
  \centering
    \includegraphics[width=\linewidth]{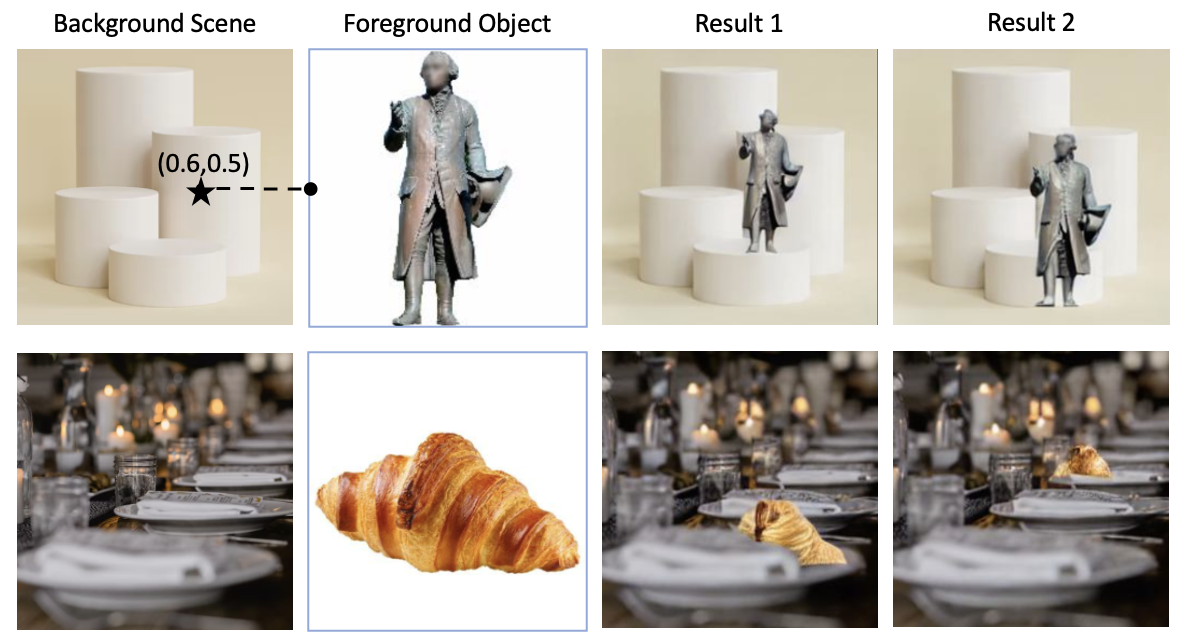}
  \caption{MADD can give different feasible solutions for ambiguous prompts such as point and blank.}
  \label{fig:diverse}
  \vspace{-1mm}
\end{figure}

In Figure~\ref{fig:affordance}, MADD adjusts a person’s position to stand on the ground rather than floating, aligns a car’s orientation with the lane, and scales coffee beans to match the scene. The model can also place objects without explicit prompts and generate diverse, reasonable insertions based on user input and learned affordance, as shown in Figure \ref{fig:diverse}.

\begin{figure}[t]
  \centering
    \includegraphics[width=0.95\linewidth]{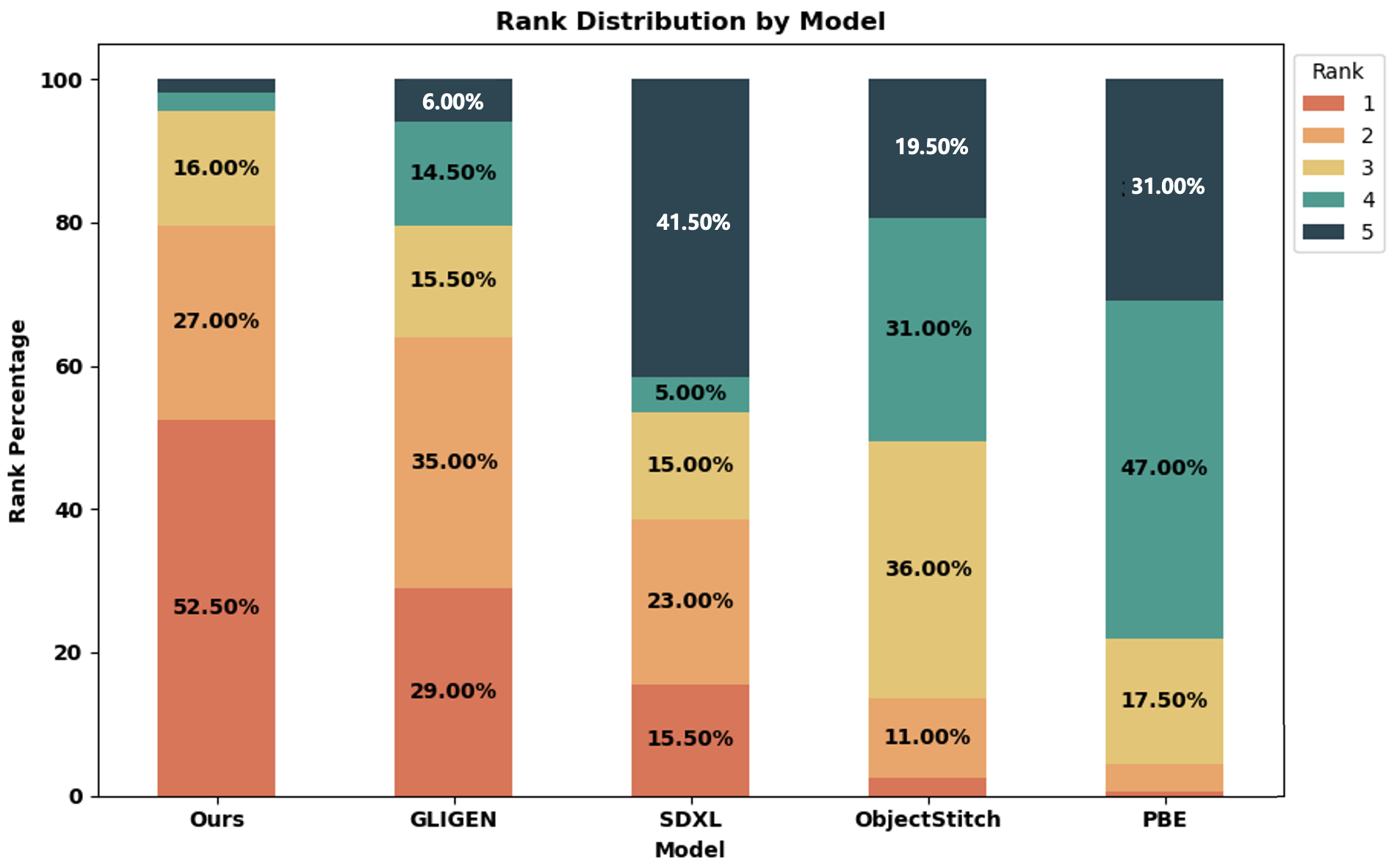}
    \vspace{-3mm}
    \caption{Rank distribution for different methods. Our method has the most proportion of rank 1 and least proportion of rank 5.}
  \label{fig:user-study-overall}
  \vspace{-4mm}
\end{figure}

\bfsection{Human Evaluation}
To test the generalization ability of our MADD model, we performed affordance insertion on in-the-wild images and compared the results with Stable Diffusion XL~\cite{podell2023sdxl}, GLI-GEN~\cite{li2023gligen}, PBE~\cite{yang2023paint} and ObjectStitch~\cite{song2023objectstitch}. Instead of merely relying on metrics like FID and CLIP Score, we also conducted user study to achieve more comprehensive comparison. We asked 10 users to rank 10 groups of composited image generated from from different model according to the following criteria: 1) Foreground and Background Integration; 2) Foreground Clarity and Detail; 3) Foreground Appearance Consistency with Reference; 4) Lighting and shade on Foreground; 5) Color Consistency. Figure~\ref{fig:user-study-overall} shows the distribution of rank for different models, where rank 1 and rank 5 represent the best and worst quality respectively. Our model achieves 50\% of Rank-1 place and 1.60\% of the Rank-5 place, which outperforms other methods. 
Please refer to supplementary for details of each criteria. Our model consistently achieved the highest proportion of Rank-1 placements across all evaluation criteria, as indicated by the largest segments in each pie chart. Especially for keeping the consistency of foreground appearance with reference image. This dominance in Rank-1 distribution across multiple criteria highlights the model’s superior performance compared to others in all aspects.

\begin{figure}[t]
  \centering
    \includegraphics[width=1.0\linewidth]{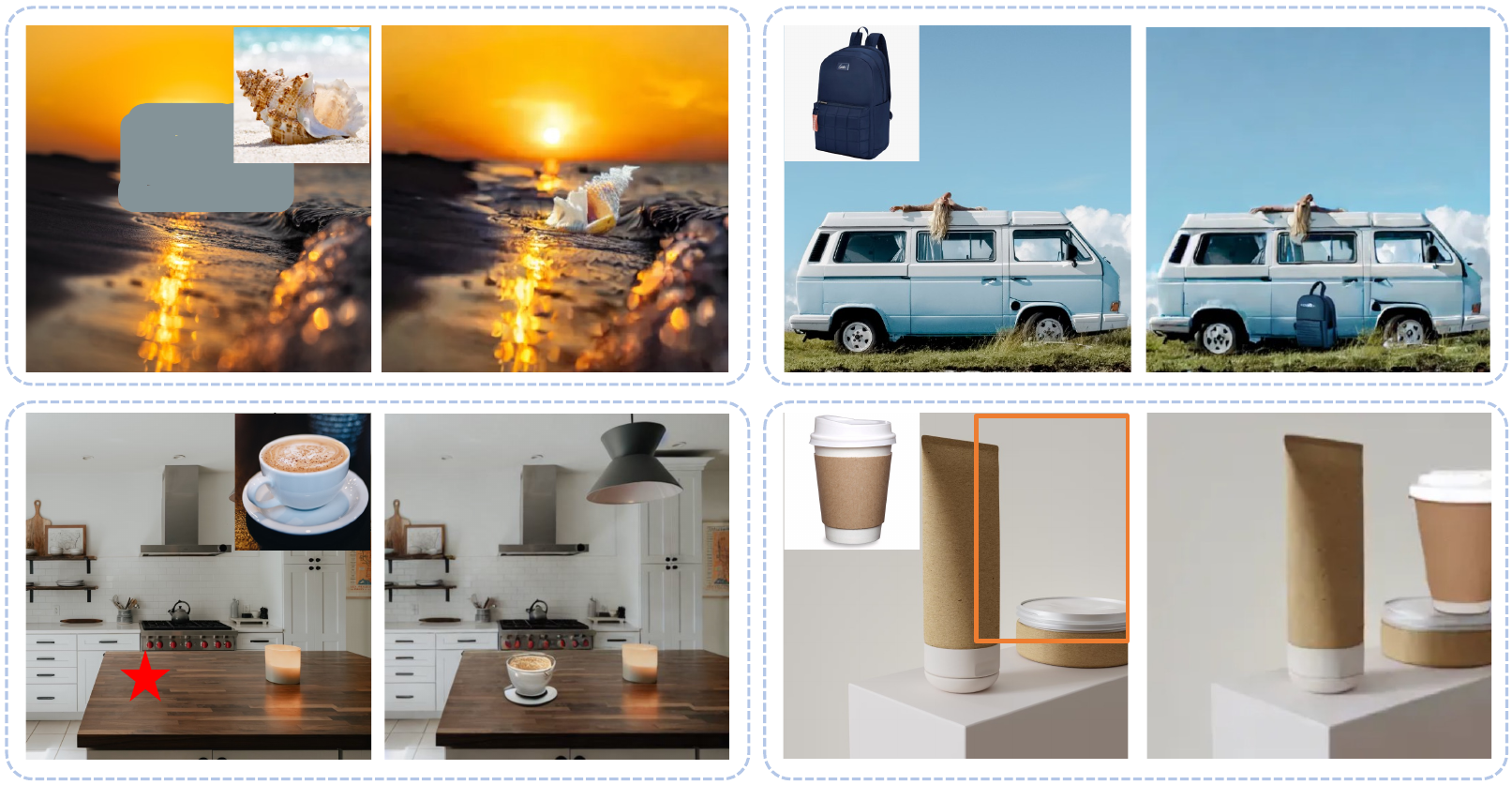}
  \caption{More in-the-wild affordance-insertion examples. The model can generate an affordance-feasible solution to insert the foreground objects according to the background scene.}
  \label{fig:moreaffordance}
  \vspace{-4mm}
\end{figure}

\bfsection{More Results}
Figure~\ref{fig:moreaffordance} presents additional in-the-wild examples of affordance-aware object insertion with various position prompts. For ambiguous prompts, such as points or null inputs, MADD infers a reasonable position and adjusts the foreground accordingly. For bounding boxes and masks, MADD treats them as guidance rather than strict constraints, refining placements to ensure coherence. Please refer to the supplementary for additional examples that further demonstrate MADD’s ability to generate suitable and diverse insertions in real-world scenarios.

\section{Conclusion}

In this paper, we extend the concept of affordance beyond human-centered tasks by introducing affordance-aware object insertion, enabling seamless object placement into scenes while adhering to affordance principles. To support this, we construct SAM-FB, a dataset of 3.16 million foreground-background pairs spanning 3,000+ object categories. We propose MADD, a dual-stream diffusion model that jointly denoises foreground appearance and object masks, leveraging affordance relationships for context-aware insertion. MADD accommodates various position prompts and can infer placements autonomously. Our model generalizes well across diverse objects and outperforms prior diffusion-based composition methods, achieving coherent and affordance-aware insertions.
\newpage

{
    \small
    \bibliographystyle{ieeenat_fullname}
    \bibliography{main}
}
\appendix
\clearpage
\setcounter{page}{1}
\maketitlesupplementary

\section{More Implementation Details}

\subsection{SAM-FB Dataset} 
\begin{figure*}[t]
  \centering
  \begin{subfigure}{0.44\linewidth}
    \includegraphics[width=0.9\linewidth]{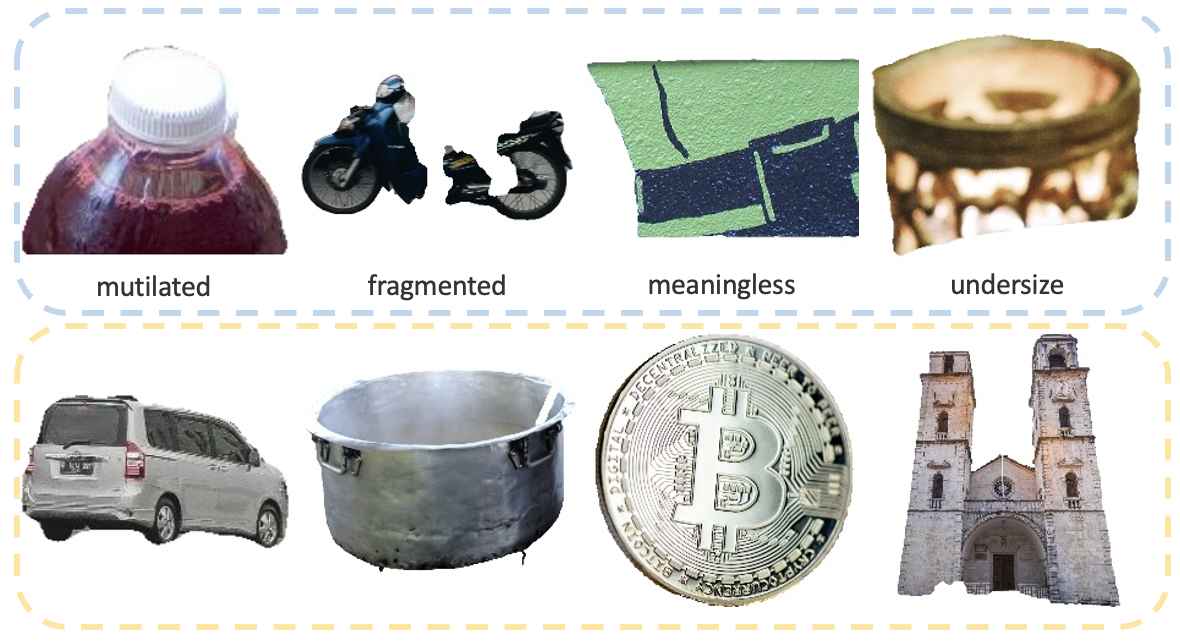}
    \caption{Examples for foreground quality control}
    \label{fig:sample}
  \end{subfigure}
  \hfill
  \begin{subfigure}{0.44\linewidth}
    \includegraphics[width=0.9\linewidth]{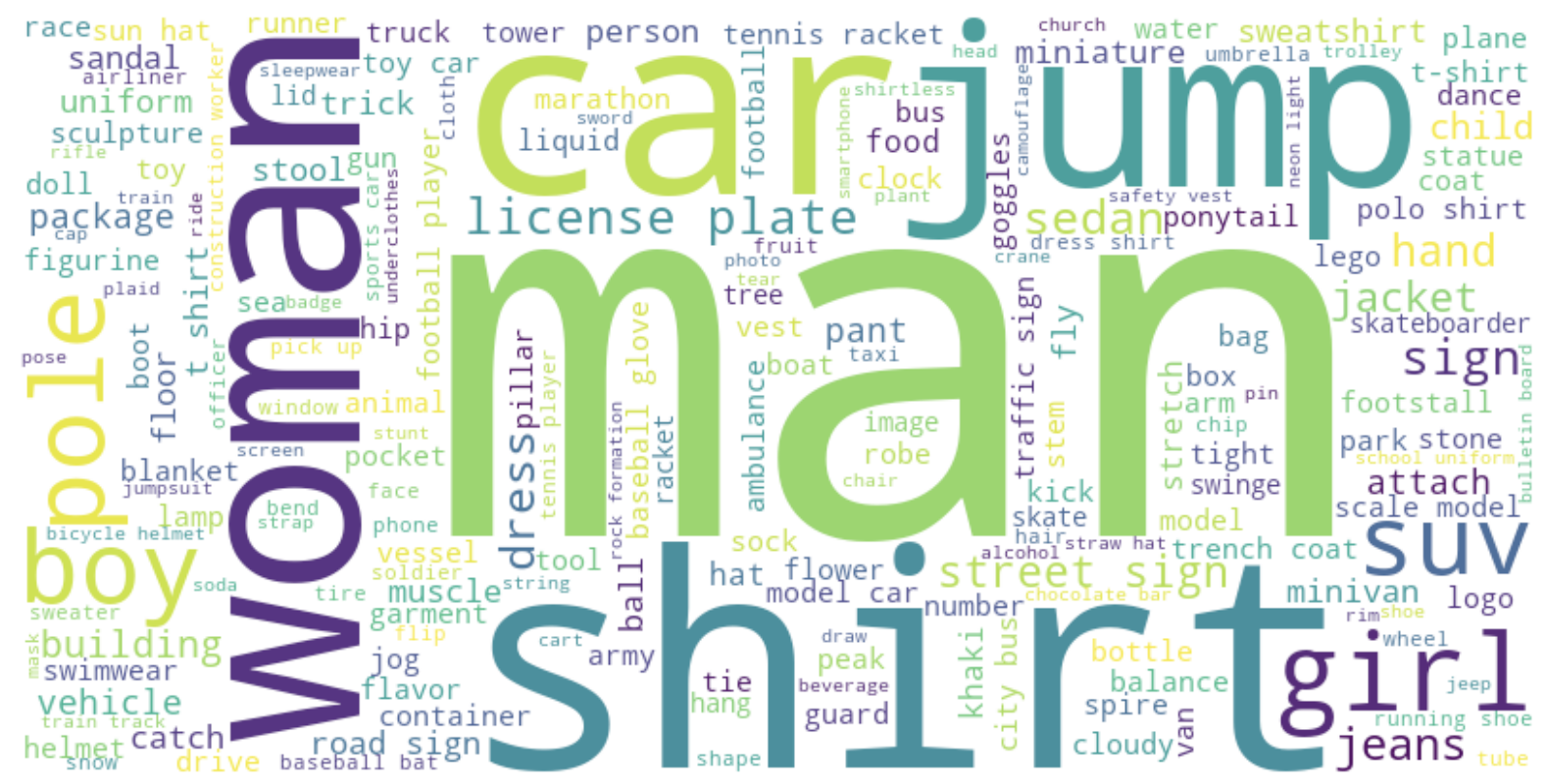}
    \caption{Word cloud of foreground categories}
    \label{fig:category}
  \end{subfigure}
 \caption{\ref{fig:sample} shows the candidate foreground samples in the pipeline. The upper row shows four low-quality samples. The lower row shows the samples after data quality control. \ref{fig:category} shows the word cloud of foreground categories in the SAM-FB dataset.}
  \label{fig:app-dataset}
\end{figure*}

Figure \ref{fig:sample} shows the object masks before and after the data quality control. We see that with our designed data quality control, the foreground object masks have better quality. Figure \ref{fig:category} illustrates the word cloud of our SAM-FB dataset, we observe that our SAM-FB dataset consists of diverse object categories.

\subsection{Traning Details}
We constructed our dual-stream UNet based on the Stable Diffusion Inpainting v1.5 model, incorporating several modifications. Specifically, we replicated both the first down-sampling block (including \verb|conv_in| and the first \verb|Down Block|) and the last up-sampling block (including the last \verb|Up Block| and \verb|conv_out|) to accommodate dual-stream inputs and outputs. These independent blocks serve as the expertise input-output branch. Skip connection is also performed between the corresponding expertise input and output branches. To bypass the need to start from scratch, we initialized our model using the pre-trained Stable Diffusion Inpainting v1.5 checkpoint for the unchanged blocks.
Fine-tuning is resource-intensive. To optimize our computing resources, we employed a gradual scaling-up approach to train our model. Initially, the model was trained at a resolution of \(128\times128\), with a batch size of 1024 and a learning rate of \(1.25 \times 10^{-4}\) for 35K steps on 2 A-100 GPUs. This phase included 5,000 warm-up steps, followed by a constant schedule. We then fine-tuned the model at a higher resolution of \(256\times256\), reducing the batch size to 256 and adjusting the learning rate to \(5 \times 10^{-5}\). 
For the diffusion process, the time step is 1000 with a linear noise scheduler. 
We use 2048 samples of SAM-FB for testing and the rest for training. 
For each foreground object, all different position prompts $\boldsymbol{p}$ can be generated with the mask $\boldsymbol{s}$. We apply classifier-free guidance~\cite{ho2022classifier}, dropping all conditions with 0.1 probability.
To ensure a fair comparison with other methods, we re-implemented and re-trained the closest method, Human Affordance, since the authors did not release either the model weights or the full dataset they used. We adopted the diffusion model architecture provided by the authors and trained it using our SAM-FB dataset. Additionally, we replaced the mask input in the original model with our position map to ensure compatibility with the SAM-FB dataset.

\subsection{Data Augmentation}
To prevent the model from learning a copy-and-paste process, we introduce different augmentations for the background images, foreground images, and position prompts.

\noindent\textbf{Background Augmentation}
Given that the images in the SA-1B dataset are not square, we rescale and crop them focusing on the objects. Our first step is to resize each source image so that its shorter edge measures 256 pixels. Following this, we randomly center a \(256\times256\) bounding box around each valid object mask, randomly chosen to provide varied backgrounds for the same object. This technique introduces slight background differences for identical objects. Additionally, we allow the bounding boxes to partially crop the objects. This strategy is specifically designed to enable the model to learn from scenarios where objects are partially obscured by the edges of the image.



\noindent\textbf{Foreground Augmentation}
To prevent the model from simply copying and pasting foreground objects and to increase the diversity of foreground objects, data augmentation is necessary. We used geometric and color augmentations based on Kulal's work~\cite{kulal2023putting} and StyleGAN-ADA~\cite{karras2020training}. Geometric augmentations included isotropic scaling, rotation, anisotropic scaling, and cutout, each with a probability of 0.4, 0.4, 0.2, and 0.2 respectively. Color augmentations included brightness, contrast, saturation, image-space filtering, and additive noise, each with a probability of 0.2.

\begin{table*}[t]
\small
\centering
\begin{tabularx}{0.95\linewidth}{@{\hspace{1mm}}p{3.70cm}p{0.8cm}<{\centering}p{0.8cm}<{\centering}p{0.8cm}<{\centering}p{0.8cm}<{\centering}p{0.8cm}<{\centering}p{0.1mm}<{\centering}p{0.8cm}<{\centering}p{0.8cm}<{\centering}p{0.8cm}<{\centering}p{0.8cm}
<{\centering}p{0.8cm}<{\centering}}
\toprule
& \multicolumn{5}{c}{\textbf{FID}$\downarrow$} & & \multicolumn{5}{c}{\textbf{CLIP Score}}\\
\cmidrule{2-6} \cmidrule{8-12}
\vspace{-0.55cm}\hspace{0.8cm}\textbf{Method} & mask & bbox & point & null & Avg. & & mask & bbox & point & null & Avg.\\
\midrule
Baseline& 16.05 & 17.31 & 18.73 & 21.18 & 18.32 && 0.8209 & 0.8254 &0.7768&0.7509&0.7935\\
+Dual Diffusion & 13.91 & 14.12 & 14.58 & 14.92 & 14.38 &&0.8582& 0.8578 &0.8435&\underline{0.8032} & \underline{0.8407} \\
+Expertise branch& \textbf{13.53} & \textbf{13.60} & \textbf{13.66} & \textbf{13.96} & \underline{\textbf{13.69}} &&\textbf{0.8727}&\textbf{0.8658}&\textbf{0.8567}&\underline{0.8034} & \underline{\textbf{0.8497}}\\
\midrule
\end{tabularx}
\caption{Experimental results on SAM-FB test set. The difference between the four kinds of prompts indicates that the performance will be better with a more precise position prompt.}
\label{table:moreablation}
\end{table*}

\noindent\textbf{Position Prompt Augmentation}
Our method requires the model to perceptively adjust the position of the inserted object in response to an ambiguous position prompt; therefore, it is also necessary to augment the position prompts. For points, we perform random jittering to deviate them from their original positions. For bounding boxes, we randomly enlarge each box. We adopt mask enlarging and feathering for mask prompts.

\begin{figure*}[t]
  \centering
    \includegraphics[width=1.0\linewidth]{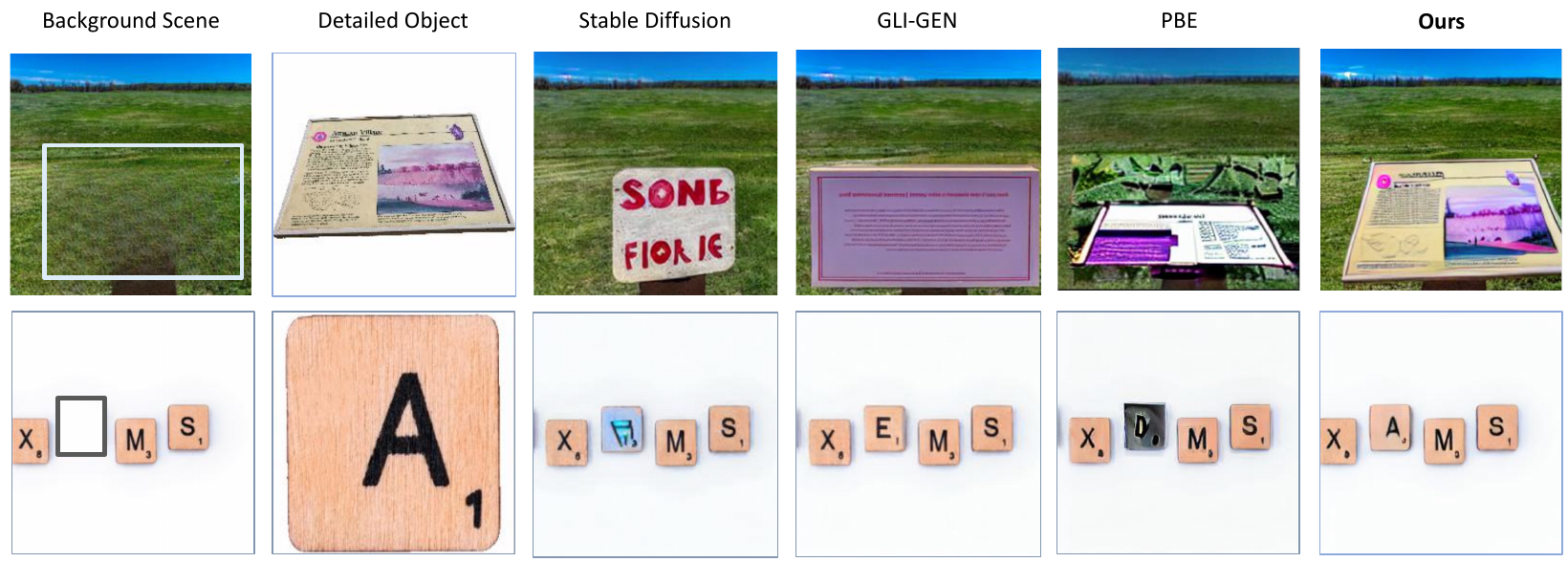}
  \caption{Example of objects with details. Our model could keep the appearance better even with some details compared with SD~\cite{rombach2022high}, GLI-GEN~\cite{li2023gligen} and PBE~\cite{yang2023paint}. The first row demonstrates the ability to keep some image texture, and the second row illustrates the ability to keep text texture.}
  \label{fig:detial}
\end{figure*}

\section{Evaluation}
\subsection{Evaluation metrics} 
Our method aims to establish a reasonable relationship between foreground objects and background scenes while maintaining the appearance of the object similar to a reference image and generating a high-quality synthetic image. To assess these two capabilities, we employ two metrics. Firstly, we use the FID score, which is widely used to measure the harmony of images obtained by generative models, to evaluate the quality of synthetic images. We use a pre-trained Inception model to extract features and calculate the FID score between generated images and ground truth images on the SAM-FB test set. Secondly, we evaluate the appearance of the inserted object and the reference foreground image using the CLIP score. 
We use a CLIP image encoder to extract features. The CLIP score is calculated as the cosine similarity between the two features. 

\begin{figure}[t]
  \centering
    \includegraphics[width=\linewidth]{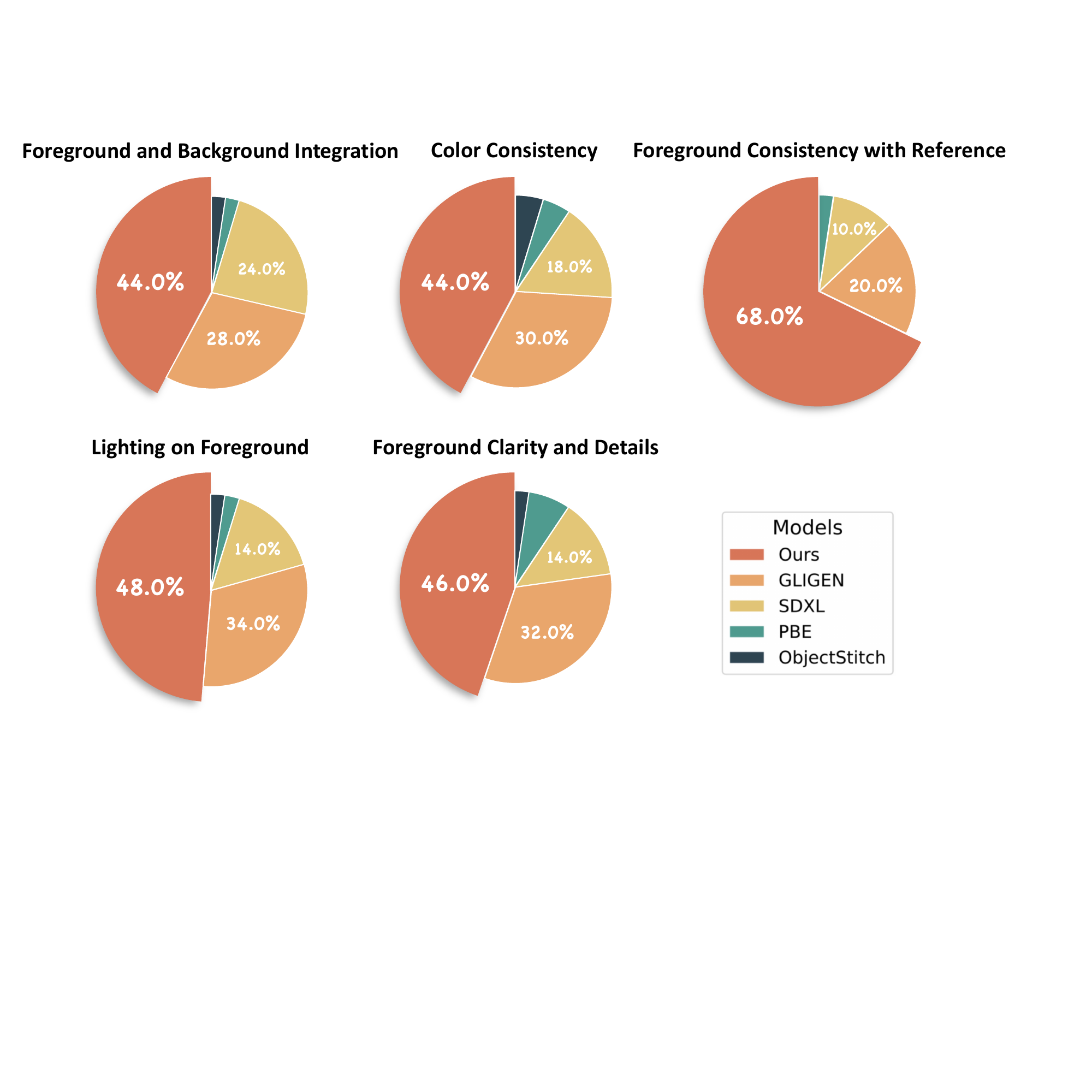}
    \caption{Rank-1 distribution for each criterion. Each pie chart represents the proportion of times each model achieved Rank-1 for a specific evaluation criterion. Our method dominates every metric.}
    \label{fig:user-study-criteria}
  \vspace{-4mm}
\end{figure}

\subsection{Human Evaluation}
When comparing the results generated by our methods and baseline methods, we asked user to evaluate the quality of the generated image following these five criteria:

\begin{itemize}
    \item \textbf{Foreground and Background Integration}: How naturally does the inserted foreground blend with the background? Does it look out of place or integrate seamlessly?
    \item \textbf{Foreground Clarity and Detail}: Assess the clarity, detail, and resolution of the \textbf{inserted foreground}.
    \item \textbf{Foreground Appearance Consistency with Reference}: Check if the inserted foreground’s appearance (shape, texture) matches the foreground in the reference image.
    \item \textbf{Lighting and shade on Foreground}: Evaluate whether highlights and shades on the inserted foreground are realistic and consistent with the background lighting.
    \item \textbf{Color Consistency}: Assess the overall color harmony. Do the inserted foreground and background tones, hues, and saturation levels align?
\end{itemize}

We conducted our user survey by asking 10 users to rank outputs from different model in 10 affordance insertion settings, therefore gaining 100 data points.

\begin{figure}[htb]
   \vspace{-3mm}
  \centering
   \includegraphics[width=\linewidth]{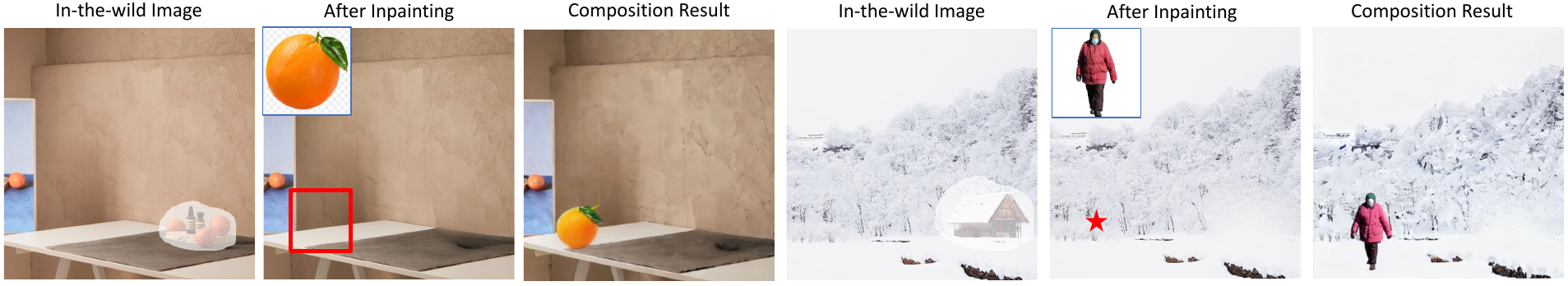}
    \vspace{-7mm}
   \caption{Affordance insertion away from inpainted region}
   \label{fig:inpaint}
   \vspace{-4mm}
\end{figure}

\subsection{Effect of Inpaintings}
Figure~\ref{fig:inpaint} shows results if the inserted location is away from masked region. 
It illustrates that the effects of inpainting artifacts are limited. 

\section{More Results}

\subsection{Details Maintaining}
Using DINOv2 features, the model preserves the appearance more effectively than other image editing models, especially for objects with detailed textures. Figure \ref{fig:detial} compares our model with other image editing models. Our model demonstrates the ability to retain the details of an object’s appearance, even texture on the object.

\subsection{SAM-FB Test Split}
\begin{figure}[t]
  \centering
    \includegraphics[width=1.0\linewidth]{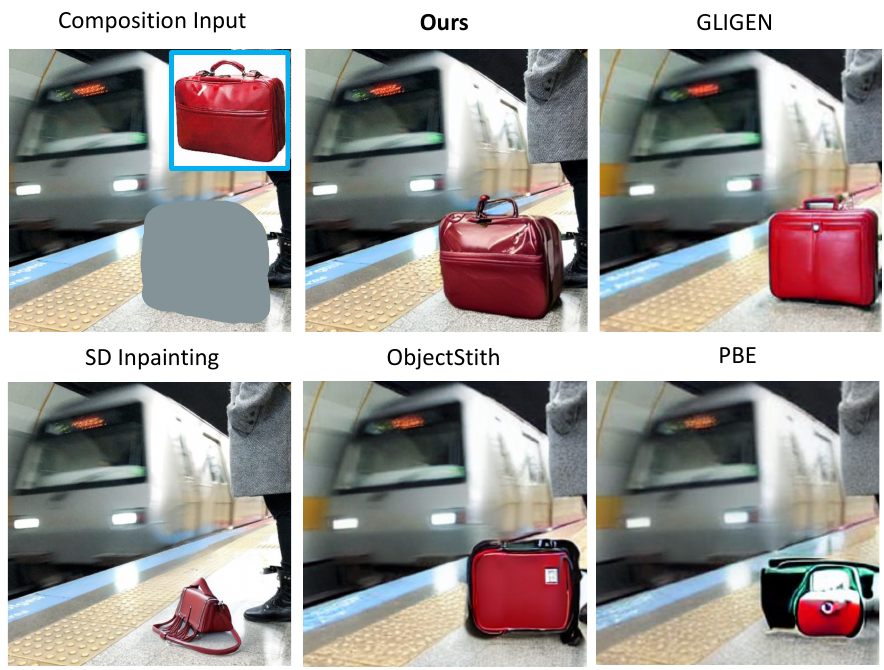}
  \caption{Samples on SAM-FB test split. Our model inserted the bag with authentic appearance.}
  \label{fig:test}
\end{figure}

Figure~\ref{fig:test} shows a visualization result with different baselines on SAM-FB test split. Our model generated more authentic results compared to other baseline models.

\subsection{Detailed Ablations}
We show the detailed ablation results on different types of position prompts and the result is presented in Table \ref{table:moreablation}. 
With all the designs, the model achieves the lowest FID score and the highest CLIP score on average.

\begin{figure*}[t]
  \centering
    \includegraphics[width=1.0\linewidth]{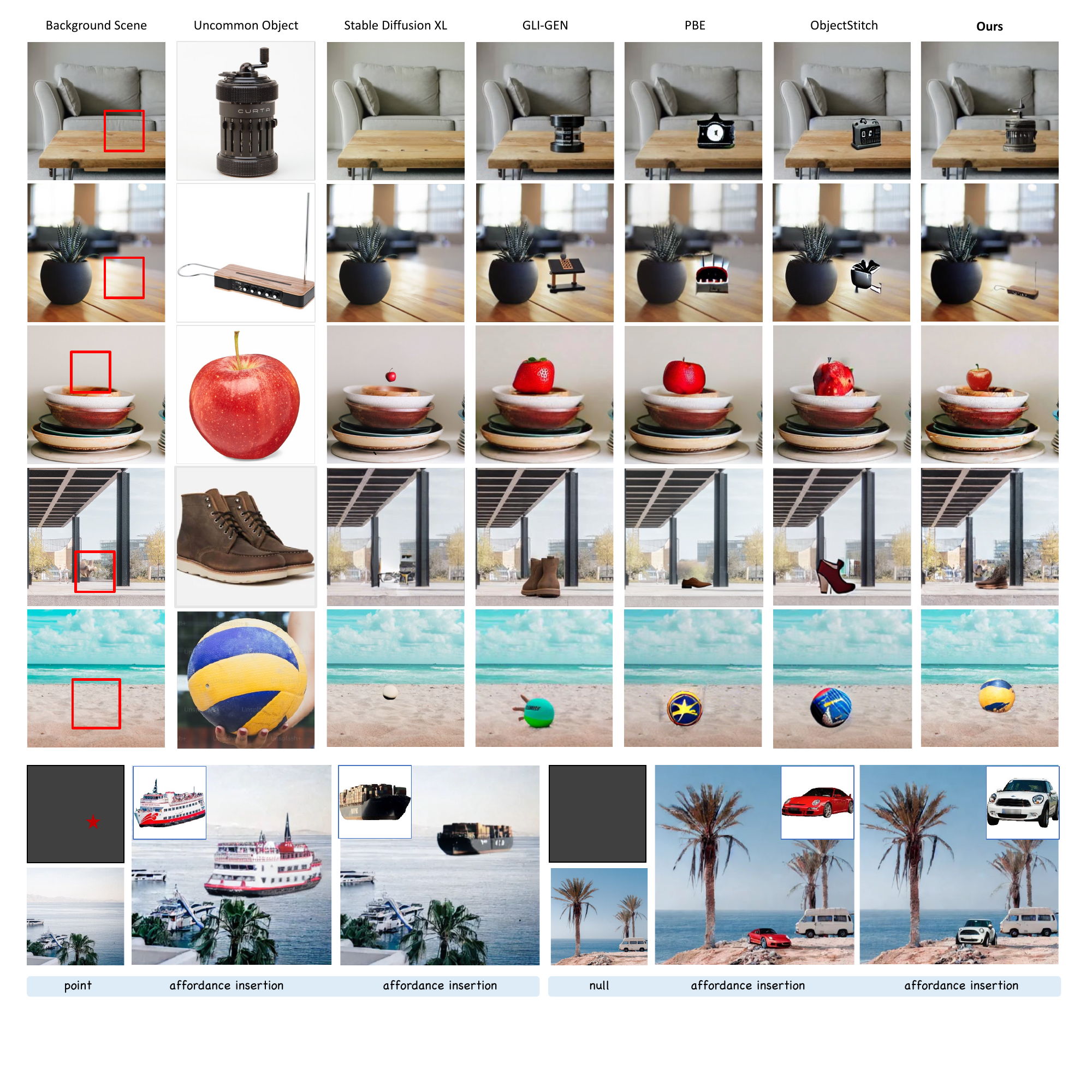}
  \caption{Example of in-the-wild insertion results with details. Our model could keep the appearance better and adjust the foreground's properties better compared with SD~\cite{rombach2022high}, GLI-GEN~\cite{li2023gligen} and PBE~\cite{yang2023paint} on common objects. In the last row, the model generated reasonable insertion when provided ambiguous prompts.}
  \label{fig:common}
\end{figure*}

\begin{table}[t]
\setlength{\tabcolsep}{3mm} 
\centering
\resizebox{0.7\linewidth}{!}{  
\begin{tabular}{lcc}
\toprule
{Training Datasets} & FID ($\downarrow$) & CLIP ($\uparrow$) \\
\midrule
COCO & 31.17 & 0.75 \\
SAM-FB & \textbf{28.45} & \textbf{0.77} \\
\bottomrule
\end{tabular}
\vspace{-2mm}
}
\caption{Ablation study results on SAM-FB test set.}
\label{table:app-ablation}\vspace{-5mm}
\end{table}

\subsection{Training Datasets}
To illustrate SAM-FB helps the model on affordance insertion task, we retrained our model on 50k COCO2014 and SAM-FB subset, each for 10k steps, then evaluated on PascalVOC as in-the-wild benchmark. Table~\ref{table:app-ablation} shows SAM-FB lead to lower FID and higher CLIP score.

\subsection{In-the-wild Generalization}
We compare our methods with other baseline models on in-the-wild images, and Figure~\ref{fig:common} shows the visualization results for comparison. In the first five rows, we performed affordance insertion task providing a bounding box on both common objects like apple and uncommon objects like cruta. Our methods generated the images with highest quality, authentic to the reference foreground with proper lighting. In the last row, we show more affordance insertion results when providing ambiguous prompts. The model will automatically find the proper affordance relationship and adjust the location and view of the foreground object. It's notable that in the first case, when inserting the cruise, it is actually the view from the back of the reference image. 

\begin{table}[t]
\setlength{\tabcolsep}{3mm} 
\centering
\begin{tabular}{l|cc}
     \hline
         Method &  Anydoor &  Ours  \\
         \hline
         CLIP Score ($\uparrow$) & 0.8209 &  \textbf{0.8658}\\
          DINO Score ($\uparrow$) & 0.667 & \textbf{0.693} \\
          RfHE($\downarrow$) & \underline{2.18} & \underline{2.21} \\
          \hline
\end{tabular}
\vspace{-1.1em}
\caption{Anydoor comparison} \label{tab:comp-anydoor}
\vspace{-2mm}
\end{table}

\begin{figure}[htb]
   \vspace{-3mm}
  \centering
   \includegraphics[width=\linewidth]{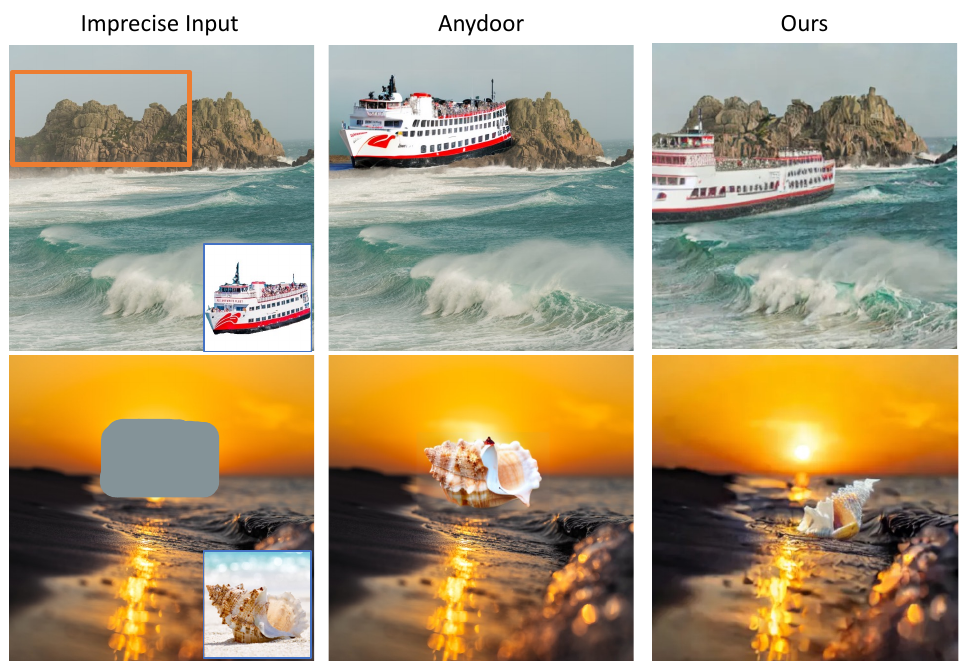}
    \vspace{-7mm}
   \caption{Affordance insertion compare with Anydoor}
   \label{fig:affordance-insertion}
   \vspace{-4mm}
\end{figure}

\subsection{Comparison with Anydoor}
Table~\ref{tab:comp-anydoor} reports quantitative results on SAM-FB and Rank from Human Evaluation (RfHE) on in-the-wild images from web (\textit{Unsplash}). Users prefer MADD for more reasonable insertion (Figure~\ref{fig:affordance-insertion}). Anydoor introduces artifacts around objects.

\subsection{Video Demo}
Please refer to the video demo in the attachment.

\subsection{Failure Cases}

\begin{figure}[t]
  \centering
    \includegraphics[width=1.0\linewidth]{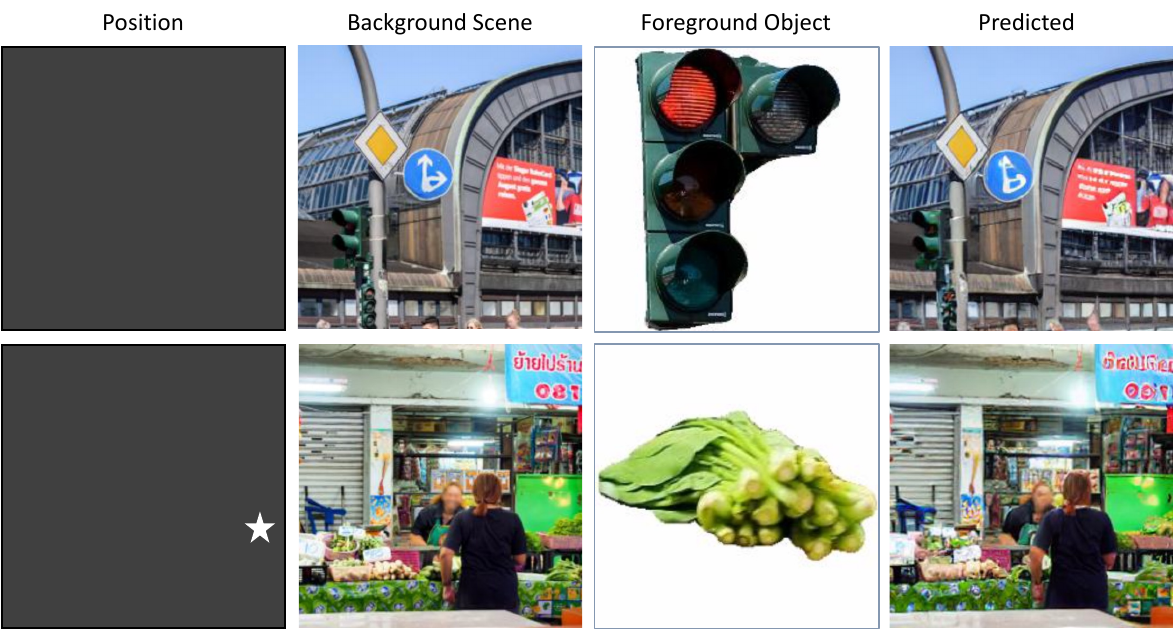}
  \caption{Some failure cases when using our model to perform affordance insertion.}
  \label{fig:fail}
\end{figure}

Figure \ref{fig:fail} shows some failure cases when using the model to perform affordance insertion. Generally, when prompted null position, it requires the model to search for a possible position to insert the object. However, if there are already similar objects in the scene \eg, traffic light in the first row, it is easy to mislead the model and end up with inserting nothing. When the background is too complex and the foreground object is too small, such as a vegetable in a supermarket shown in the second row, it is also difficult for the model to insert the object correctly. 

\end{document}